%% file: main.tex
\documentclass[letterpaper]{article} 
\usepackage[submission]{aaai25}  
\usepackage{times}  
\usepackage{helvet}  
\usepackage{courier}  
\usepackage[hyphens]{url}  
\usepackage{graphicx} 
\usepackage{natbib}  
\usepackage{caption} 
\usepackage{algorithm}
\usepackage{algorithmic}

\usepackage{newfloat}
\usepackage{listings}
\DeclareCaptionStyle{ruled}{labelfont=normalfont,labelsep=colon,strut=off} 
\floatstyle{ruled}
\newfloat{listing}{tb}{lst}{}
\floatname{listing}{Listing}
\title{OPTDTALS: Approximate Logic Synthesis via Optimal Decision Trees Approach}

\author{
    Hao Hu, Shaowei Cai\footnote{Corresponding Author}
}
\affiliations{
    Key Laboratory of System Software (Chinese Academy of Sciences) and State Key Laboratory of Computer Science, \\
    Institute of Software, Chinese Academy of Sciences, Beijing, China\\
    \{huhao, caisw\}@ios.ac.cn
}

\usepackage{bibentry}

\usepackage{amsmath, mathtools, amsthm}
\usepackage{algorithm, algorithmic}
\usepackage{float}
\usepackage{graphicx}
\usepackage{booktabs}
\usepackage{tikz}
\usetikzlibrary{shapes}
\usetikzlibrary{positioning}
\usetikzlibrary{arrows}
\usetikzlibrary{fit}
\tikzset{
    textellipsnode/.style={
        draw, ellipse, minimum width=10mm, fill=black!10
    },
    textellipsnodenofill/.style={
        draw, ellipse, minimum width=10mm
    },
    textplainnode/.style={
        ellipse, minimum width=10mm
    },
    textsquarenode/.style = {
        draw, rectangle, minimum width=8mm
    },
    graphroundnode/.style = {
        draw, circle, minimum width=5mm, fill=white
    },
    graphsquarednode/.style = {
        rectangle, draw, minimum width=6mm, fill=black!10 
    },
    graphnonode/.style = {
        circle, minimum width=5mm, fill=white
    }
}

\usepackage{array}
\newcolumntype{H}{>{\setbox0=\hbox\bgroup}c<{\egroup}@{}}
\usepackage{multicol, multirow, colortbl}

\begin{document}

\maketitle

\input{paper_structure/_1_symbols}
\input{paper_structure/0_abstract}
\input{paper_structure/1_introduction}

\input{paper_structure/2_technical_background}

\input{paper_structure/3_related_works}
\input{paper_structure/4_OPTDTALS_approach}

\input{paper_structure/5_experiments}

\input{paper_structure/6_conclusion}
\bibliography{reference}

\end{document}

%% file: paper_structure/_1_symbols.tex
\def\boolvarspace{\mathbf{X}}
\def\aboolvar#1{x_{#1}}

\def\boolfunc{f}
\def\boolfuncappro{\tilde{f}}

\def\distance#1#2{d(#1, #2)}

\def\mdepth#1{md_{#1}}
\def\depth#1{d_{#1}}

\def\circuit{Cir}
\def\circuitapprox{\Tilde{\circuit{}}}
\def\circuitsubstitute#1#2{Cir(#1 \rightarrow #2)}
\def\subcircuit#1{s_{#1}}
\def\subcircuitapprox#1#2{SA_{#1, #2}}

\def\errthreshold{err}
\def\step{step}

\def\after#1{\textcolor{blue}{#1}}
\def\before#1{\textcolor{lightgray}{#1}}

%% file: paper_structure/0_abstract.tex
\begin{abstract}

{The growing interest in Explainable Artificial Intelligence (XAI) motivates promising studies of computing optimal Interpretable Machine Learning models, especially decision trees.
Such models generally provide optimality in compact size or empirical accuracy.
{Recent works focus on improving efficiency due to the natural scalability issue.
The application of such models to practical problems is quite limited.}
As an emerging problem in circuit design,
Approximate Logic Synthesis (ALS) aims to reduce circuit complexity by sacrificing correctness.
Recently, multiple heuristic machine learning methods have been applied in ALS, which learns approximated circuits from samples of input-output pairs.}

{In this paper, we propose a new ALS methodology realizing the approximation via learning optimal decision trees in empirical accuracy.
Compared to previous heuristic ALS methods, the guarantee of optimality achieves a more controllable trade-off between circuit complexity and accuracy.
Experimental results show clear improvements in our methodology in the quality of approximated designs (circuit complexity and accuracy) compared to the state-of-the-art approaches.}


\end{abstract}

%% file: paper_structure/1_introduction.tex
\section{Introduction}

By providing structures that are inherently understandable by humans,
as a classical interpretable machine learning (ML) model, 
decision trees obtain increasing concerns in explainable artificial intelligence (XAI)~\cite{Rudin-J-NATMI19-StopBlackBox, Rudin-Corr21-InterpretableML}.
Compared to traditional greedy decision tree models inducing accurate but large trees~\cite{Breiman-Chapter84-CART, Quinlan-J-ML86-InductionDT}, recent exact methods focus on finding \textit{optimal} decision trees of a metric~\cite{Ignatiev-IJCAI21-InterpretableMLSummary, Costa-J-AIR23-RecentAdvanceDT}.
There are mainly three different metrics to optimize: \textit{tree size}, \textit{tree depth}, and \textit{empirical accuracy}.
The exact methods optimizing \textit{tree size} (respectively, \textit{tree depth}) target to find decision trees of the smallest \textit{tree size} (respectively, \textit{tree depth}) with perfect empirical accuracy.
Some typical methods include~\cite{Bessiere-CP09-MinimisingDTSize, Narodytska-IJCAI18-SATDT} (optimal tree size),
~\cite{Avellaneda-AAAI20-SATDTDepth, Janota-SAT20-SATDTPath, Alos-Corr21-ODTMaxSAT-depth, Schidler-AAAI21-SATDTLargeDataSets} (optimal tree depth).
To optimize \textit{empirical accuracy}, the exact methods aim to find topology-restricted decision trees with the highest empirical accuracy, where the restriction is mostly the tree depth
~\cite{Nijssen-JDMKD10-DL8, Bertsimas-JML17-OCTMIP, Hu-NIPS19-OSDT, Verwer-AAAI19-BinOCT, Verhaeghe-JConstrains20-CPDT, Aglin-AAAI20-DL8.5, Hu-IJCAI20-MaxSATDT, Shati-CP21-SATDTNonBinaryFeature, Demirovic-JMLR22-MurTree, Hu-AAAI22-MaxSATBDD, Shati-CP23-SATBDDForClassification, Demirovic-ICML23-Blossom, Huisman-AAAI24-OPTSurvivalTrees}.



As an emerging paradigm in the modern Electronic Design Automation (EDA) 
{process}, Approximate Logic Synthesis (ALS) offers benefits in terms of circuit area, power, or latency, by relaxing the requirement of full accuracy.
During the last decade, multiple ALS methods have been proposed~\cite{Scarabottolo-JPIEEE20-ALSSurvey},
where most of them focus on functional approximation.
That is, to simplify the function implemented by a circuit in the corresponding gate-level netlist.
There are three major approaches corresponding to different abstract levels: \textit{netlist transformation}, \textit{logic rewriting}, and \textit{approximate high-level synthesis}.
The \textit{netlist transformation} approach is realized by removing some electrical nodes or substituting some wires with others
~\cite{Shin-DATE11-NT-SAF, Venkataramani-DATE13-NT-SASIMI, Liu-ICCAD17-NT-SCALS, Schlachter-J-TVLSI17-NT-GLP, Scarabottolo-DATE18-NT-CircuitCarving}.
The \textit{logic rewriting} approach
acts on the function truth table, which modifies the values of outputs for a subset of inputs to achieve the approximation~\cite{Yang-DAC00-LR-ALSBDD, Mishchenko-DAC06-LR-LRAIG, Venkataramani-DAC12-LR-SALSA, Miao-ICCAD13-LR-ALSgeneralErrorMagntitude, Wu-DAC16-LR-Anefficient-multilevelALS, Hashemi-DAC18-LR-BLASYS-Orig, Hashemi-DATE19-LR-MatrixFactorization, Ma-J-TCAD22-LR-BLASYS-Journal}.
The \textit{approximate high-level synthesis} approach targets 
the behavioral level, 
which modifies a portion of behavioral operations~\cite{Nepal-DATE14-AHLS-ABACUS, Lee-DATE17-AHLS-AHLSC, Mrazek-DAC19-AHLS-AutoAX, Zeng-ICCAD21-LR-ALSDTSHAP}.

Due to the close relationship between ML and ALS, recently, in the IWLS Contest 2020, multiple supervised ML methods have been explored to realize ALS~\cite{Rai-DATE21-IWLS21}.
Those ML-based ALS methods learn unknown Boolean functions from the sampled input-output pairs of several circuits, which are classified as \textit{logic rewriting} approaches.
In brief, recent works in this field majorly apply 
{heuristic} decision trees 
{for profiting} 
from the convenience of converting the 
{tree structures} learned into sets of rules, 
where rules are suitable for generating the circuit
~\cite{Zeng-ICCAD21-LR-ALSDTSHAP, Abreu-ISCAS21-LR-FastLODT, Huang-J-TCAD23-LR-CircuitLearningDTtoDG}.

Inspired by the usage of heuristic decision trees in ALS, 
{in this paper,
we devise a new methodology called \textbf{OPT}timal \textbf{D}ecision \textbf{T}rees-based \textbf{A}pproximate \textbf{L}ogic \textbf{S}ynthesis (\textbf{OPTDTALS}), that applies optimal decision trees in \textit{empirical accuracy} to ALS.
To our best knowledge, OPTDTALS is the first attempt at offering approximated circuit designs with the guarantee of optimality.}
{Considering the restricted tree topology measures the circuit complexity, the optimality in accuracy provides a more controllable trade-off between accuracy and circuit complexity compared to previous ALS methods.}
To scale 
{OPTDTALS} to large circuits, 
we propose to first partition an input circuit into multiple manageable sub-circuits, 
then apply 
{OPTDTALS} to those 
sub-circuits.
We introduce a design space exploration heuristic to replace the sub-circuits in the final approximated circuit iteratively.
Our experimental results demonstrate clear improvements in the quality of approximated designs comparing the proposed method to the state-of-the-art 
ALS method.

As an NP-hard problem~\cite{Hyafil-J-IPL76-ODTNP},
computing optimal decision trees is time-consuming.
We emphasize the significance of OPTDTALS, which provides more compact approximated circuits than existing ALS methods with the guarantee of optimality.
Meanwhile, it is a novel applicable approach for 
optimal decision tree algorithms into the circuit design domain.
The rest of this paper is organized as follows.
We first introduce preliminaries,
Then, we will overview some closely relevant previous works. 
Next, 
we present the framework of our novel methodology proposed.
Finally, We provide comprehensive experimental results. 

%% file: paper_structure/2_technical_background.tex
\section{Technical Background}
\label{sec:technical_background}



\subsection{Quality-of-Result Metric}

Multiple error metrics are used in the quality-of-result (QoR) phase to evaluate the quality of the approximated circuit.
This paper considers monitoring the \textit{Average Relative Error}.
For a circuit with input space $\boolvarspace{}$, the \textit{Average Relative Error} of the approximated Boolean function $\boolfuncappro{}$ comparing to the original Boolean function $\boolfunc{}$ is defined as follows:

\begin{equation*}
    \frac{1}{|\boolvarspace{}|}\sum_{\aboolvar{} \in \boolvarspace{}}\frac{\distance{\boolfunc{(\aboolvar{})}}{ \boolfuncappro{(\aboolvar{})}}}{\lVert \boolfunc{(\aboolvar{})} \rVert}
\end{equation*}

where $\distance{\boolfunc{(\aboolvar{})}}{\boolfuncappro{(\aboolvar{})}}$ indicates the difference between the exact and approximated outputs of the same input $\aboolvar{}$.
Generally, \textit{Hamming Distance} is employed to measure the difference by counting the number of bit flips in $\boolfuncappro{(\aboolvar{})}$ with respect to $\boolfunc{(\aboolvar{})}$, which is defined as follows:

\begin{equation*}
    \distance{\boolfunc{(\aboolvar{})}}{\boolfuncappro{(\aboolvar{})}} = \lVert \boolfunc{(\aboolvar{})} - \boolfuncappro{(\aboolvar{})} \rVert
\end{equation*}

In the rest of the paper, we use the QoR metric to represent the \textit{Average Relative Error}.

\subsection{Decision Trees}

As a classical interpretable ML model, decision trees are commonly used to make classifications.
A decision tree contains a root node, several branching nodes, and leaf nodes.
For each branching node (or the root node), a feature is selected as the test for the judgment. 
Each outcoming edge of the branching node (or the root node) corresponds to the case of the feature selected.
The leaf node corresponds to a class.
Each path from the root node to a leaf node is associated with a series of judgments.
To predict an unseen sample, find the corresponding path from the root node to a leaf node based on its values for the series of tests, the prediction is the class associated with the leaf node. 

From the view of ALS, a binary decision tree is employed to represent a Boolean function, 
where a binary feature corresponds to an input bit and the binary class relates to the output bit.
The sampled input-output pairs are used as the training set.
When all possible input-output pairs of the input space are sampled, it is evident that maximizing \textit{empirical accuracy} of the decision tree is equivalent to minimizing the \textit{Average Relative Error} metric of ALS.


%% file: paper_structure/3_related_works.tex
\section{Related Works}
\label{sec:related_works}



{This section introduces further details for some classical \textit{logic rewriting} ALS methods and some optimal decision tree approaches in empirical accuracy.} 

As a representative method using Boolean Optimization, SALSA~\cite{Venkataramani-DAC12-LR-SALSA} constructs a QoR circuit of a single output by combining the 
{original} and approximated circuits.
{The {positive} output of the QoR circuit indicates the approximated circuit is acceptable for a maximum error bound.}
Therefore, the logic of the approximated circuit is free to simplify as long as the output of the QoR circuit stays as a tautology.
SALSA computes the \textit{observability don't cares} for each output of the approximated circuit to minimize the circuit.
{Another ALS method }BLASYS~\cite{Hashemi-DAC18-LR-BLASYS-Orig, Ma-J-TCAD22-LR-BLASYS-Journal} captures the truth table of the input circuit as a Boolean matrix $\textbf{A}$, then apply a 
{Boolean Matrix Factorization (BMF)} algorithm to decompose $\textbf{A}$ into two matrices $\mathbf{B}$ and $\mathbf{C}$, such that $\mathbf{A} \approx \mathbf{B}\mathbf{C}$ with minimum error.
The approximated circuit is created by combining the circuits synthesizing for $\mathbf{B}$ and $\mathbf{C}$.
BLASYS applies heuristic ASSO BMF Algorithm~\cite{Miettinen-KDD11-ASSO}.
Furthermore,~\cite{Yang-DAC00-LR-ALSBDD} proposes to generate a binary decision diagram representing all potential approximations
with a maximum error bound, then find the minimal cover with the smallest size as the final approximated design.
Regarding the efficiency and relevance to our method, we consider BLASYS the state-of-the-art ALS method.

There are two strategies to compute optimal decision trees in empirical accuracy: one is to encode the ML problem into a standard declarative combinatorial optimization problem and then solve it via general solvers, like MaxSAT approaches~\cite{Hu-IJCAI20-MaxSATDT, Shati-CP21-SATDTNonBinaryFeature}, CP approach~\cite{Verhaeghe-JConstrains20-CPDT}, etc.
The other is to extract solutions directly from the search space via the dynamic programming approach, like DL8, DL8.5~\cite{Nijssen-JDMKD10-DL8, Aglin-AAAI20-DL8.5}, MurTree~\cite{Demirovic-JMLR22-MurTree}, and Blossom~\cite{Demirovic-ICML23-Blossom}. 
These algorithms scale well to large datasets by leveraging branch independence as the subtrees can be optimized independently.
Regarding time efficiency, we apply dynamic programming approaches to OPTDTALS, such as DL8.5 and Blossom.


%% file: paper_structure/4_OPTDTALS_approach.tex
\begin{figure*}[hbt!]
    \centering
    \includegraphics
    [width=\linewidth]
    {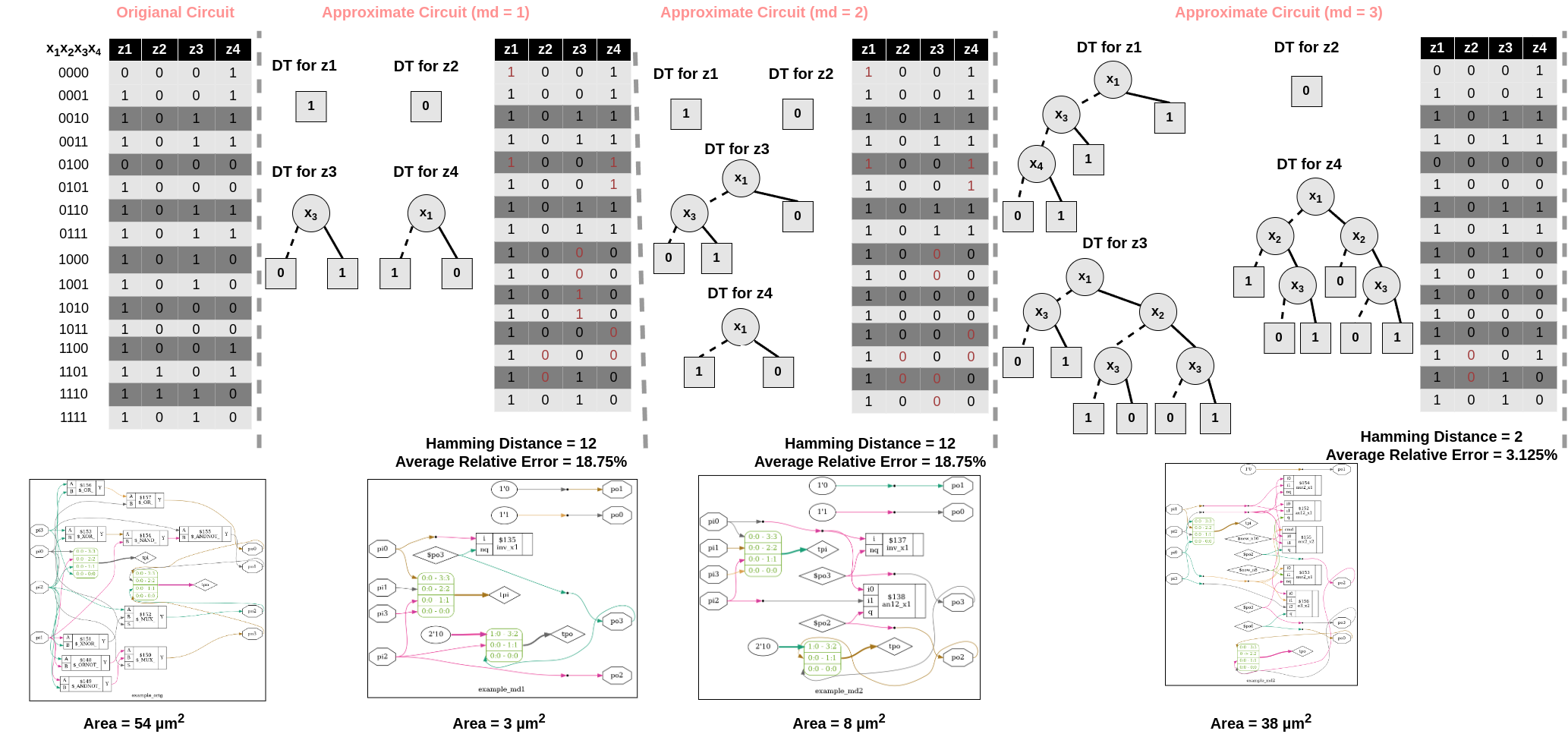}
    \caption{An Illustrative Example of OPTDTALS using DL8.5 algorithm to learn optimal decision trees with various maximum depths. The wrong predictions are marked in red. Circuits are synthesized via Yosys using the open ssxlib013 Liberty.}
    \label{fig:4_1_illustrative_example}
\end{figure*}

\section{Proposed Methodology}
\label{sec:proposed_method}

\subsection{Circuit Approximation Via 
{Optimal Decision Trees}}


In our proposed approach, OPTDTALS, we consider the ALS process as an ML process of learning 
{optimal decision trees in empirical accuracy.}
{For instance, to handle a single-output logic circuit with $k$ inputs,
a dataset of size $2^{k}$ consisting of the pairs of all inputs of the input space and their corresponding output is first generated.
Then, OPTDTALS learns an optimal decision tree
for this dataset.
The predictions 
for all inputs of the input space form the approximated outputs,
{which is used to be} further synthesized into the gate-level circuit.
OPTDTALS is easy to extend for a multi-output case by considering it as multiple 
{independent} single-output cases. 
In brief, for a general
multi-output logic circuit of 
$m$ outputs, OPTDTALS learns $m$ optimal decision trees, each of them corresponding to a separate output bit.}





{With the guarantee of the optimality in empirical accuracy, 
for OPTDTALS, 
the maximum tree depth reflects the approximation level 
{as it directly measures the approximated circuit complexity.}
In general, 
{deeper} optimal decision trees 
lead to better prediction performance but more complicated structures.
Figure~\ref{fig:4_1_illustrative_example} 
{shows} an illustrative example of $4$-input, $4$-output arbitrary logic circuit.
We present the truth table of the original circuit 
{and the synthesized design with Yosys Tool~\cite{Yosys}
~\footnote{https://yosyshq.net/yosys/about.html} using the open ssxlib013 Liberty~\footnote{https://www.vlsitechnology.org/html/ssx\_description.html}.}
We then provide the approximated circuits generated by 
{OPTDTALS} using the DL8.5 algorithm
with 
maximum depths from $1$ to $3$.
The optimal decision trees of each output bit, 
their average relative error, and their corresponding synthesized circuits are shown in Figure.
The wrong predictions are marked in red.
Compared to the original design, the approximated design 
{of the maximum depth $3$} reduces $29.6\%$ circuit area while compromising only $3.125\%$ in circuit accuracy. 
Meanwhile, with the decrease of the maximum depth, 
{OPTDTALS} offers more aggressive approximated designs in reducing circuit area by sacrificing more accuracy.
In this example, 
when the maximum depth is $1$, we can reduce 
{$94.4\%$ circuit area} 
while sacrificing $18.75\%$ circuit accuracy.

Compared to ALS methods proposed in the IWLS Contest 2020,
our method can handle multi-output circuits, and consider the full input space instead of a subset sampled via Monte Carlo.
Moreover, our method provides the guarantee of optimality in circuit accuracy.
However, the main challenge of our approach is the scalability problem as the truth table grows exponentially with the increase of circuit inputs.

\subsection{Scaling Up for Large Circuits}
\label{sec:proposed_method-large_circuit}

Learning optimal 
{decision trees} naturally suffers the scalability issue.
Even the most time-efficient dynamic programming approaches are unable to 
{handle training sets of size more than 20,000 examples.}
Therefore, 
{OPTDTALS} is acceptable for circuits of maximum $14$ inputs, as $2^{14} = 16384$.

To scale 
{OPTDTALS} to large circuits, similar to BLASYS,
we propose first to partition a given circuit into several manageable sub-circuits, 
then approximate those sub-circuits via 
{OPTDTALS}. 
We apply KahyPar~\cite{Schlag-ALENEX16-Partition-KwayHyperPar}
recursively until all sub-circuits have maximum $k$ inputs and maximum $m$ outputs.
The number of $k$ and $m$ are preset parameters determined by the budget of the computing resource.
In this paper, we 
{apply} $k=14$ and $m=5$.
The selection of $5$ outputs accounts for limiting the number of optimal decision trees.

{We evaluate the QoR metric of the whole circuit instead of sub-circuits in isolation, 
as a small error in the sub-circuit can propagate leading to large errors.}
We use $\circuitsubstitute{\subcircuit{i}}{\subcircuitapprox{\subcircuit{i}}{\mdepth{i}}}$ to represent the approximated circuit by substituting the original sub-circuit $\subcircuit{i}$ with its corresponding approximation $\subcircuitapprox{\subcircuit{i}}{\mdepth{i}}$ generated by OPTDTALS, 
where $\mdepth{i}$ is the maximum depth of the optimal decision tree.
As it is infeasible to evaluate the entire approximated circuit with the full input space,
we use Monte Carlo sampling to estimate the QoR metric.
{We mention that, for all sub-circuits, the full input space is sampled  when applying OPTDTALS.}

\input{Algos/Algo_procedure_subcircuit}
\input{Algos/Algo_OPTIMALALS_Partitioned}

After 
{the large circuit is partitioned,}
we need to decide the approximation level (the maximum depth) of each sub-circuit, 
then identify a good order of 
{substituting} 
the sub-circuits with the approximated designs so that the circuit 
{complexity} decreases fast with the loss of accuracy.
Therefore, we apply a similar design space exploration heuristic in BLASYS
to replace the sub-circuits iteratively. 
Algorithm~\ref{alg:optimalals_par} describes the 
{iterative} 
process.
{The vector recording the maximum depth of OPTDTALS to explore for each sub-circuit is called \textit{max\_depth stream},
{which is initialized with values of $\mdepth{}^{0}$}.
With the the initial maximum depth 
$\mdepth{}^{0}$, we obtain the approximated design for each sub-circuit by calling the $ApproxSubCir$ function described in Algorithm~\ref{alg:optimalals_subcircuit}.
The \textit{max\_depth stream} is updated by the real depths of the optimal decision trees learned.}
Then, Lines $7$-$20$ details how the search process works:
substitute the sub-circuit iteratively with the smallest loss until the QoR metric exceeds the error threshold.
The loss 
{in Line 9} measures the trade-off between the circuit area and accuracy,
{which is smaller when more benefit in area reduction is achieved with less sacrifice in accuracy.}
The \textit{max\_depth stream} would be updated with the 
{search} step size preset, and the new corresponding approximated sub-circuit would be generated.

Furthermore, 
{to avoid possible local optima,}
instead of replacing the sub-circuit with the smallest loss in each iteration,
{we substitute several sub-circuits with the top smallest losses separately.
Then we can explore multiple search paths via monitoring multiple different \textit{max\_depth streams}, each representing a replacement of different sub-circuit.}

%% file: Algos/Algo_procedure_subcircuit.tex
\begin{algorithm}[!h]
    \caption{\label{alg:optimalals_subcircuit}Function $ApproxSubCir$}

  \begin{algorithmic}[1]
    \INPUT{Sub-circuit: $\subcircuit{i}$, Maximum Depth: $\mdepth{}$, 
    {Optimal Decision Tree Algorithm: OPTDT}\\}
    \OUTPUT{Approximated Sub-circuit: $\subcircuitapprox{\subcircuit{i}}{\mdepth{i}}$}
  
    \FUNCTION{$ApproxSubCir(\subcircuit{i}, \mdepth{}, {\text{OPTDT})}$}
        \STATE $A$ = Construct truth table of $\subcircuit{i}$
        
        \STATE \COMMENT{Assume $\subcircuit{i}$ has $m_i$ outputs, learn $m_i$ optimal decision trees via given OPTDT method}

        \FOR{$ind = 1$ to $m_i$}
            \STATE \COMMENT{$A[ind]$ indicates the $ind$-th column, $\mdepth{}^{ind}$ indicates the real tree depth}
            \STATE $[OPT_{ind}$, $\mdepth{}^{ind}]$ = {OPTDT}($A[ind]$, $\mdepth{}$)
        \ENDFOR

        \STATE \COMMENT{$\mdepth{i}$ indicates the value in max\_depth stream}
        \IF{all $OPT_{ind}$ have no error}
            \STATE $\mdepth{i} = \min_{ind}\mdepth{}^{ind}$
        \ELSE
            \STATE $\mdepth{i} = \mdepth{}$
        \ENDIF

        \STATE $\subcircuitapprox{\subcircuit{i}}{\mdepth{i}} =$ Circuit {synthesized} by all $OPT_{ind}$

    \STATE \RETURN{$\subcircuitapprox{\subcircuit{i}}{\mdepth{i}}$}
    \ENDFUNCTION
  \end{algorithmic}

\end{algorithm}

%% file: Algos/Algo_OPTIMALALS_Partitioned.tex
\begin{algorithm}[!h]
    \caption{\label{alg:optimalals_par}{OPTDTALS}: Approximate Logic Synthesis via Optimal Decision Trees 
    for Large Circuits}

  \begin{algorithmic}[1]
    \INPUT{Input Original Circuit: $\circuit{}$, Error Threshold: $\errthreshold{}$, Initial Maximum Depth: $\mdepth{}^{0}$, Optimal Decision Tree method: OPTDT, 
    Iteration Step Size: $\step{}$\\}
    \OUTPUT{Approximated Circuit: $\circuitapprox{}$}

    \STATE subcircuits = Decompose $\circuit{}$ via KahyPar recursively


    \FOR{each $\subcircuit{i} \in$ subcircuits}

        \STATE $\subcircuitapprox{\subcircuit{i}}{\mdepth{i}} = ApproxSubCir(\subcircuit{i}, \mdepth{}^{0}, {\text{OPTDT}})$
         
    \ENDFOR


    \STATE $\circuitapprox{} = \circuit{}$

    \WHILE{$QoR(\circuitapprox{}) \leq \errthreshold{}$}
        \FOR{each $\subcircuit{i} \in $ subcircuits with $\mdepth{i} > 1$}
            \STATE $\circuitapprox{}^{'} = \circuitsubstitute{\subcircuit{i}}{\subcircuitapprox{\subcircuit{i}}{\mdepth{i}}}$

            \STATE $loss_i = (area(\circuitapprox{}^{'}) - area(\circuit{})) / QoR(\circuitapprox{}^{'})$
        \ENDFOR

        \STATE \COMMENT{Substitute the sub-circuit with the smallest loss}
        \STATE $g = \arg \min_i (loss_i)$
        \STATE $\circuitapprox{} = \circuitsubstitute{\subcircuit{g}}{\subcircuitapprox{\subcircuit{g}}{\mdepth{g}}}$

        \STATE $\mdepth{g} = \mdepth{g} - \step{}$


        \IF{$\mdepth{g} > 1$}
            \STATE $\subcircuitapprox{\subcircuit{g}}{\mdepth{g}} = ApproxSubCir(\subcircuit{g}, \mdepth{g}, {\text{OPTDT}})$ 
        \ENDIF
        
    \ENDWHILE

    \STATE \RETURN{$\circuitapprox{}$}
  \end{algorithmic}

\end{algorithm}

%% file: paper_structure/5_experiments.tex
\section{Experimental Results}
\label{sec:experiments}

\input{Tables/5_1_IWLS_average_results}
\input{Figures/5_1_IWLS_Results_combined}

In this section, we perform two evaluations for 
{OPTDTALS}
in terms of the accuracy and complexity of approximated designs generated.
The first evaluation is performed on the IWLS Contest 2020 benchmarks, where we compare 
{OPTDTALS} 
with the ML-based ALS method proposed in the contest.
Another evaluation is performed on several classical combinatorial circuits from the ISCAS85 benchmarks~\cite{Hansen-J-DTC99-ISCAS85Benchmarks}~\footnote{https://web.eecs.umich.edu/\textasciitilde{}jhayes/iscas.restore/benchmark.html},
where we compare 
{OPTDTALS} 
with the state-of-the-art ALS method (BLASYS).
We ran all experiments on a cluster using AMD EPYC-7763 @2.45GHz CPU and 1TB memory running Ubuntu 20.04 LTS.





\input{Tables/5_1_IWLS_benchmarks_overview}

\subsection{Evaluation in the IWLS Contest 2020 Benchmarks}

To explore the relationship between ML and ALS, IWLS Contest 2020 proposed $100$ single output benchmarks for different types of functions.
Table~\ref{tab:iwls20_benchmarks} provides an overview, 
where Case $00$ to $49$ are in \textit{arithmetic} domain, Case $50$ to $79$ are in \textit{random logic} domain, and Case $80$ to $99$ are in \textit{machine learning} domain.
For each benchmark, there are separate training, validation, and testing set. 
Each set contains $6400$ examples and is in PLA format~\cite{Espresso-Rudell-89-These}.
The approximated designs 
are synthesized into the \textit{And-Inverter Graph} (AIG)~\cite{AIGER-Biere-07} format, where the number of $AND$ gates in AIG measures the circuit complexity. 

Multiple ML-based ALS methods are proposed in the contest, while~\cite{Rai-DATE21-IWLS21} concludes that no technique dominates across all the benchmarks.
Considering the relevance, 
performance, and code availability, we select Team2's solution as the state-of-the-art ML-based ALS method to compare.
Although ranking 
the $5$-th place among all teams,
Team2's solution obtains a very close prediction performance compared to the best one.
Moreover, 
{Team2's solution is firmly relevant to our method, as it 
uses the J48 and PART classifiers 
in WEKA Tool~\cite{Hall-J-SIGKDD09-WekaTool}, which are traditional heuristic decision trees.}

{Our methodology makes it easy to assess those benchmarks by treating them as single-output cases.
Instead of generating the dataset sampling the full input space for each benchmark, 
we can directly learn an optimal decision tree using the provided training set and then evaluate the accuracy via the provided testing set.
In detail,}
we apply two different versions of 
{OPTDTALS} using DL8.5 algorithm 
(\textbf{OPTDTALS-DL8.5}) and Blossom algorithm 
(\textbf{OPTDTALS-Blossom}), where both of them are
time-efficient dynamic programming approach.
To avoid the possible overfitting, 
{we generate multiple approximated designs with different maximum depths from 2 to 10,}
then select the one with the best validation accuracy
as the final approximated design.
The total time is limited to $30$ hours.
To create a baseline of AIG sizes for benchmarks, we feed the training PLAs to ABC
\footnote{We use the recommended synthesis command in the IWLS Contest 2022: \textit{read\_pla; collapse; sop; strash; dc2; write\_aiger}} to perform the exact synthesis.

Figure~\ref{fig:5_1_IWLS_results} demonstrates the evaluation results 
{in terms of} accuracy and circuit complexity,
where different methods are marked in different colors.
In addition, Table~\ref{tab:5_1_iwls_average_results} presents detailed average experimental results 
{grouped} by the function category of benchmarks.
The column ``$AND$'' indicates the number of $AND$ gates, the column ``\textbf{Test}'' (respectively ``\textbf{Train}'' and ``\textbf{Val}'') indicates the testing accuracy (respectively to training and validation accuracy) in percent,
{and the column ``$\depth{avg}$'' indicate the average depth of optimal decision trees explored.}
The best values between different methods are marked in blue.
For accuracy, we mark those values very close to the best one ($< 1\%$) in yellow.
{We mark the average evaluation of different domains in cyan,}
and 
the global average evaluation in red.
We mention the equivalence between high testing accuracy and low QoR metric.

Several observations from Figure~\ref{fig:5_1_IWLS_results} and Table~\ref{tab:5_1_iwls_average_results} are presented as follows.
At first, the exact synthesis designs realized by ABC suffer from vast circuit complexity and the overfitting problem.
Then, regarding testing accuracy, although slight disadvantages are observed for benchmarks of \textit{random logic domain} (Case $50$-$79$), both 
{OPTDTALS versions} perform better than Team2's solution in global.
Especially, 
both {OPTDTALS versions} provide far more compact approximated designs concerning the number of $AND$ gates.
On average (Case $00$-$99$), the approximated designs found via 
{OPTDTALS versions} contain only $38\%$ (OPTDTALS-DL8.5) or $13\%$ (OPTDTALS-Blossom) number of $AND$ gates compared to the ones generated via Team2's solution. 
{We do not report the running time, {as both OPTDTALS versions fail to generate all approximated designs of maximum depths from 2 to 10 within the total time limit for almost all benchmarks.}
The other methods finish the approximation mostly in several minutes.
From the results of $\depth{avg}$ in Table~\ref{tab:5_1_iwls_average_results}, we observe that DL8.5 is more time-efficient than Blossom for these benchmarks as it explores more different designs.} 
It is clear that scalability seriously limits the performance of our methodology.

\subsection{Evaluation in the Combinatorial Circuits}

This experiment aims to compare our methodology with the state-of-the-art \textit{logic rewriting} ALS method (BLASYS) for classical combinatorial circuits.
For small circuits, we generate the truth tables and then directly learn corresponding optimal 
{decision trees.} 
For the larger ones, we apply the scaling-up process proposed before. 
The combinatorial circuits 
evaluated (Table~\ref{tab:5_2_ISCAS_benchmarks}) contain classical 8-bit adder, 7-bit multiplier, and part of the ISACAS85 benchmarks (circuits starting with C).
All are implemented in Verilog.
{For each large circuit,} as it is impractical to enumerate all test vectors to evaluate the QoR metric, 
we generate a testbench containing $10,000$ random test vectors.
For small circuits, the testbench contains all possible test vectors.

\input{Tables/5_2_ISCAS85_benchmarks}

\begin{figure}[htb]
    \centering
    \includegraphics[scale=.27]
    {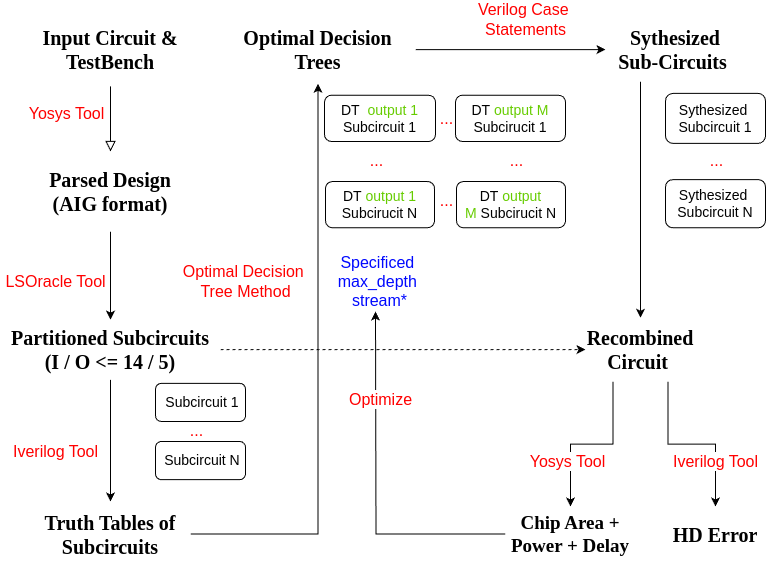}
    \caption{The general framework of OPTDTALS tool-chain.}
    \label{fig:5_2_framework}
\end{figure}

Figure~\ref{fig:5_2_framework} describes the 
{tool-chain} used in OPTDTALS.
In detail, Yosys 
parses the input circuit and assesses its circuit area with a given liberty file. 
LSOracle~\cite{Neto-ICCAD19-LSOracle}~\footnote{https://github.com/lnis-uofu/LSOracle} 
is used to partition the input circuit into multiple manageable sub-circuits with similar sizes.
For each sub-circuit, Icarus Verilog~\cite{Iverilog}~\footnote{https://steveicarus.github.io/iverilog/}  generates its corresponding truth table, which is used to learn multiple 
{optimal decision trees.}
After optimal 
{decision trees} 
for a sub-circuit are learned, we synthesize these models using the Verilog Case Statements.
Thus, an approximated design of the input circuit can be generated by partly substituting some approximated sub-circuits.
The approximated design's circuit area and QoR metric are used to iteratively optimize the \textit{max\_depth stream} as described in 
Algorithm~\ref{alg:optimalals_par}.
In this experiment, we apply 
{the DL8.5 algorithm as its time-efficiency.}
All designs are synthesized by Yosys using the open ssxlib013 liberty.

\input{Tables/5_2_results_C17}
\input{Tables/5_2_results_ISCAS_orig}
\input{Tables/5_2_results_ISCAS_approx}

\begin{figure*}[htb!]
        \centering
        \includegraphics[scale=.4]{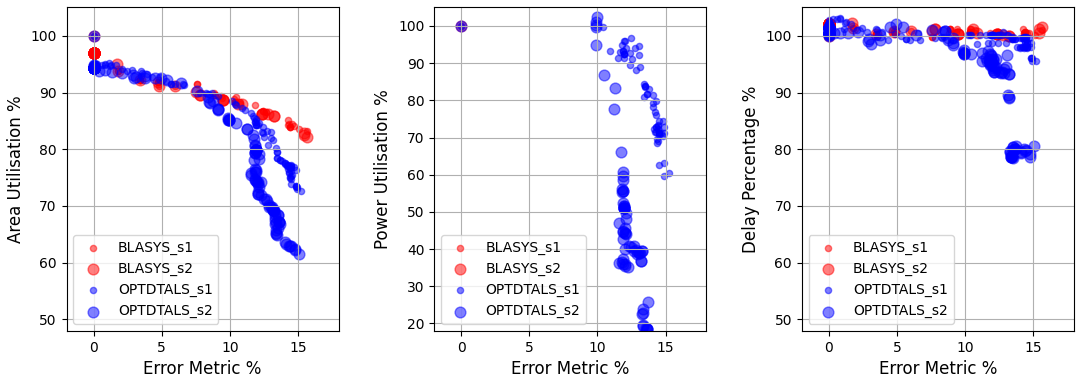}
        \caption{Tendency of area utilization (the left scatter), power (the middle one), and delay (the right one) of approximated designs of different methods using different step sizes with the increase of error for the 
        C6288 circuit.}
        \label{fig:5_2_C6288_diff_steps}
\end{figure*}


We first evaluate OPTDTALS 
on the small C17 circuit.
We consider different maximum depths for OPTDTALS-DL8.5 ($\mdepth{} \in [2,4]$) to generate different approximated designs.
Meanwhile, as the 
C17 circuit has only two outputs, BLASYS can only provide one approximated design with the factorization degree $fd$ as 1.
Table~\ref{tab:5_2_results_c17} presents the results of designs generated via different methods.
Compared to BLASYS, we observe that OPTDTALS can provide diverse
approximated designs without the limit of the number of outputs.
Meanwhile, OPTDTALS generates a design of zero areas (
{containing} no logic gates) when $\mdepth{} =1$ with the QoR metric of $25\%$.
However, a negative observation is that when $\mdepth{} = 3$, the approximated design obtained by OPTDTALS has a larger circuit area than the original but fails to 
{maintain the full correctness.}


Then we consider the general evaluation 
on the large circuits left.
The 
partitioning algorithm (KahyPar) we applied can not directly control the number of inputs and outputs of the sub-circuits generated.
Therefore, we set the initial partition number as $5$, and the KahyPar algorithm is applied recursively until all sub-circuits have a maximum of $14$ inputs and $5$ outputs.
The initial maximum depth $\mdepth{}^{0}$ to explore for each sub-circuit is set as $9$.
We preserve the top 3 best \textit{max\_depth streams} in each design space exploration iteration, and the iteration step size is set as $1$. 
For each {circuit}, 
we consider three different error thresholds ($\errthreshold{} \in \{5\%, 10\%, 15\%\}$) to end the iteration.
We report the circuit information (area, power, and delay) of the original designs in Table~\ref{tab:5_2_ISCAS_orig}.
The evaluation of 
{OPTDTALS}-DL8.5 and BLASYS for all circuits 
are summarized in Table~\ref{tab:5_2_ISCAS_approx}.
The column ``\textbf{Part}'' corresponds to the real number of sub-circuits partitioned,
the column ``\textbf{A(\%)}'' (respectively ``\textbf{P(\%)}'' and ``\textbf{D(\%)}'') represents the percentage of the approximated 
{design}
's circuit area (respectively, power and delay) compared to the original design,
and the column ``\textbf{T(s)}'' indicates the execution time in seconds.
The best values between different methods are marked in blue.
The lack of value is marked with the symbol ``-''.
In general, it occurs when no 
{better} 
approximated designs are found 
{with more sacrifice in circuit accuracy.}
A special case 
is for the 
C432 circuit, where BLASYS suffers from the \textit{logic loop} problem raised by Yosys.
From Table~\ref{tab:5_2_ISCAS_approx}, we observe that, within the same error threshold, our method shows a clear advantage over the state-of-the-art method in reducing circuit area, power, and delay.
On average for all circuits, compared to BLASYS,
{OPTDTALS-DL8.5 offers 
approximated designs minimizing $5.4\%$, $7.6\%$, and $10.9\%$ in area utilization within the $5\%$, $10\%$, and $15\%$ error thresholds.}
Furthermore, both methods find approximated designs eliminating all gates for the 
C499 with a slight loss of accuracy ($\leq 0.05\%$). 
However, 
{the report of running time indicates the scalability issue of OPTDTALS in handling large circuits.}

We {increase} the iteration step as $2$ 
{to avoid possible local optima.}
Unfortunately, except for the 
C6288 circuit, we fail to observe obvious advances. 
Figure~\ref{fig:5_2_C6288_diff_steps} demonstrates the circuit information of approximated designs of different iterations for the 
C6288 circuit.
The left scatter shows the tendency of area utilization with the increase of error,  
and the middle (right) scatter shows the power (delay).
Different methods are marked in different colors, and different step sizes are marked in different sizes.
Figure~\ref{fig:5_2_C6288_diff_steps} indicates that, for the C6288 circuit,  
{increasing the step size improves the quality of approximated designs for OPTDTALS-DL8.5.} 
Considering the 
C6288 circuit is the largest circuit evaluated, the step size may be a concerning parameter in 
{avoiding local optima for complicated circuits.}

%% file: Tables/5_1_IWLS_average_results.tex
\begin{table*}[hbt!]
    \centering
    \renewcommand{\arraystretch}{1.2}
    \setlength{\tabcolsep}{4.5pt}
    \footnotesize{}
    \begin{tabular}{ccHHcccccccccccccc} \hline
    \multirow{2}{*}{{\textbf{Cases}}} & 
    \multicolumn{3}{c}{{\textbf{ABC}}} &
    \multicolumn{4}{c}{{\textbf{Team2}}} &
    \multicolumn{5}{c}{\textbf{OPTDTALS-DL8.5 - 30h}} &
    \multicolumn{5}{c}{\textbf{OPTDTALS-Blossom - 30h}} \\
    
    \cmidrule(lr){2-4}\cmidrule(lr){5-8}\cmidrule(lr){9-13}\cmidrule(lr){14-18} 
    & \textbf{$AND$} & \textbf{Test} & \textbf{Val} & \textbf{$AND$} & \textbf{Test} & \textbf{Train} & \textbf{Val} & \textbf{$AND$} & \textbf{Test} & \textbf{Train} & \textbf{Val} & \textbf{$\depth{avg}$} & \textbf{$AND$} & \textbf{Test} & \textbf{Train} & \textbf{Val} & \textbf{$\depth{avg}$}  \\ \hline

   00-09 & 182813.80 & 50.39 & 50.01 & 151.30 & 79.84 & 84.25 & 79.85 & 23.40 & \cellcolor{yellow!40}87.84 & \cellcolor{blue!40}89.61 & \cellcolor{blue!40}87.88 & \cellcolor{blue!40}5.2 & \cellcolor{blue!40}17.00 & \cellcolor{blue!40}87.88 & \cellcolor{yellow!40}88.88 & \cellcolor{yellow!40}87.73 & 4.8 \\
   
10-19 & 158413.30 & 57.33 & 57.47 & 133.90 & 83.91 & \cellcolor{blue!40}87.26 & 84.30 & 49.10 & \cellcolor{blue!40}85.49 & \cellcolor{yellow!40}86.88 & \cellcolor{blue!40}85.55 & \cellcolor{blue!40}5.4
& \cellcolor{blue!40}22.90 & 83.49 & 84.92 & 83.90 & 4.7 \\

20-29 & 90023.50 & 51.10 & 50.91 & 643.80 & 70.17 & 81.52 & 73.72 & 385.20 & \cellcolor{blue!40}76.09 & \cellcolor{blue!40}83.66 & \cellcolor{blue!40}76.37 & \cellcolor{blue!40}5.0
& \cellcolor{blue!40}58.00 & \cellcolor{yellow!40}75.77 & 80.40 & \cellcolor{yellow!40}76.19 & 4.5\\

30-39 & 101852.50 & 49.67 & 49.74 & 35.30 & \cellcolor{blue!40}98.07 & \cellcolor{blue!40}98.63 & \cellcolor{blue!40}98.10 & \cellcolor{blue!40}12.40 & 93.10 & 93.64 & 93.23 & \cellcolor{blue!40}5.2 &
13.50 & 92.25 & 92.69 & 92.30 & 4.8 \\

40-49 & 66858.10 & 55.68 & 55.48 & 816.70 & \cellcolor{blue!40}63.49 & \cellcolor{blue!40}79.27 & \cellcolor{blue!40}65.59 & 202.80 & 62.23 & 70.46 & 62.95 & \cellcolor{blue!40}5.9 & 
\cellcolor{blue!40}129.00 & 62.06 & 68.08 & 62.71 & 5.4\\

\cellcolor{cyan!50}00-49 & 119992.24 & 52.83 & 52.72 & 356.20 & 79.10 & \cellcolor{blue!40}86.18 & \cellcolor{yellow!40}80.31 & 134.58 & \cellcolor{blue!40}80.95 & 84.85 & \cellcolor{blue!40}81.19 & \cellcolor{blue!40}5.34  & \cellcolor{blue!40}48.08 & \cellcolor{yellow!40}80.29 & 82.99 & \cellcolor{yellow!40}80.57 & 4.84\\ \bottomrule[2pt]

\hline
50-59 & 63140.40 & 64.14 & 64.16 & 144.90 & 90.04 & \cellcolor{blue!40}93.75 & 89.91 & 45.00 & \cellcolor{blue!40}91.70 & \cellcolor{yellow!40}92.90 & \cellcolor{blue!40}91.56 & \cellcolor{blue!40}6.5 & \cellcolor{blue!40}25.10 & \cellcolor{yellow!40}90.80 & 91.55 & \cellcolor{yellow!40}90.65 & 5.9 \\

60-69 & 30868.00 & 54.57 & 54.42 & 82.00 & \cellcolor{yellow!40}97.44 & \cellcolor{blue!40}98.52 & \cellcolor{yellow!40}97.32 & 29.40 & \cellcolor{blue!40}97.80 & \cellcolor{yellow!40}98.00 & \cellcolor{blue!40}97.77 & \cellcolor{blue!40}7.1 & \cellcolor{blue!40}24.40 & 96.79 & 96.98 & 96.75 & 6.5 \\

70-79 & 10203.80 & 49.28 & 49.68 & 1144.10 & \cellcolor{blue!40}86.19 & \cellcolor{blue!40}93.99 & \cellcolor{blue!40}93.63 & 479.40 & 81.15 & 87.66 & 81.31 & \cellcolor{blue!40}6.4 & \cellcolor{blue!40}98.90 & 80.83 & 84.76 & 80.91 & 5.8\\

\cellcolor{cyan!50}50-79 & 34737.40 & 56.00 & 56.09 & 457.00 & \cellcolor{blue!40}91.22 & \cellcolor{blue!40}95.42 & \cellcolor{blue!40}93.62 & 184.60 & \cellcolor{yellow!40}90.22 & 92.85 & 90.21 & \cellcolor{blue!40}6.67 & \cellcolor{blue!40}49.47 & 89.48 & 91.09 & 89.43 & 6.07 \\ \bottomrule[2pt]

80-89 & 62886.10 & 50.56 & 50.78 & 158.70 & 90.57 & 92.17 & 90.79 & 46.40 & \cellcolor{blue!40}92.46 & \cellcolor{blue!40}94.51 & \cellcolor{blue!40}92.80 & \cellcolor{blue!40}5.0 &  \cellcolor{blue!40}43.00 & \cellcolor{yellow!40}91.93 & \cellcolor{yellow!40}94.05 & \cellcolor{yellow!40}92.37 & 4.9  \\

90-99 & 519068.20 & 50.21 & 50.25 & 43.70 & 59.67 & 61.42 & 59.60 & \cellcolor{blue!40}8.40 & \cellcolor{yellow!40}64.69 & \cellcolor{blue!40}67.31 & \cellcolor{blue!40}64.46 & \cellcolor{blue!40}3.0 &  \cellcolor{blue!40}8.40 & \cellcolor{blue!40}64.71 & \cellcolor{blue!40}67.31 & \cellcolor{yellow!40}64.44 & \cellcolor{blue!40}3.0 \\ 

\cellcolor{cyan!50}80-99 & 290977.15 & 50.39 & 50.52 & 101.20 & 75.12 & 76.79 & 75.20 & 27.40 & \cellcolor{blue!40}78.57 & \cellcolor{blue!40}80.91 & \cellcolor{blue!40}78.63 & \cellcolor{blue!40}4.0 & \cellcolor{blue!40}25.70 & \cellcolor{yellow!40}78.32 & \cellcolor{yellow!40}80.68 & \cellcolor{yellow!40}78.41 & 3.95 \\

\toprule[2pt] \cellcolor{red!50}00-99 & 128612.77 & 53.29 & 53.29 & 335.44 & 81.94 & \cellcolor{blue!40}87.08 & \cellcolor{yellow!40}83.28 & 128.15 & \cellcolor{blue!40}83.25 & \cellcolor{yellow!40}86.46 & \cellcolor{blue!40}83.39 & \cellcolor{blue!40}5.47 & \cellcolor{blue!40}44.02 & \cellcolor{yellow!40}82.65 & 84.96 & \cellcolor{yellow!40}82.80 & 5.03\\ \bottomrule[2pt]
    
    \end{tabular}
    \caption{Average results of different methods in the IWLS Contest 2020 benchmarks categorized by functions. Case $00$-$49$ indicate \textit{arithmetic} domain, Case $50$ -$79$ indicate \textit{random logic} domain, and Case $80$-$99$ indicate \textit{machine learning} domain.}
    \label{tab:5_1_iwls_average_results}
\end{table*}

%% file: Figures/5_1_IWLS_Results_combined.tex
\begin{figure*}[hbt!]
        \begin{minipage}{.495\linewidth}
            \centering
            \includegraphics[scale=.45]{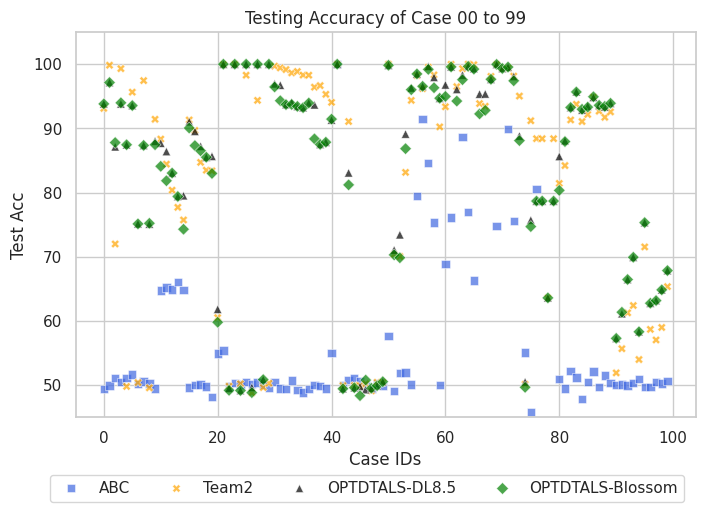}
        \end{minipage}\hfill
        \begin{minipage}{.495\linewidth}
            \centering
            \includegraphics[scale=.45]{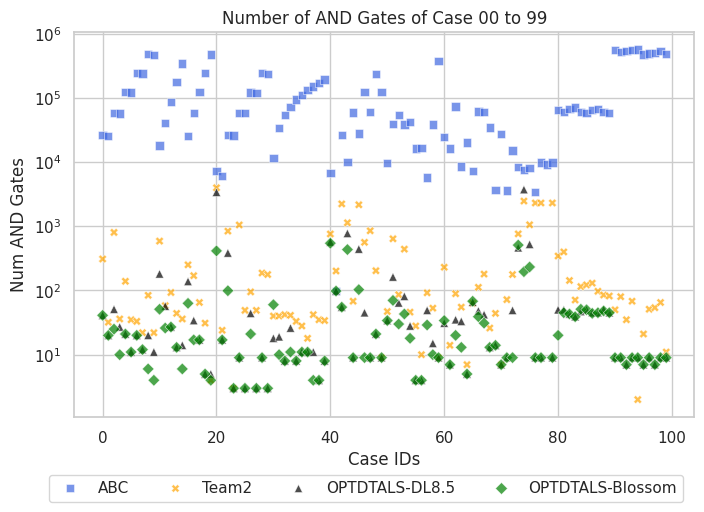}
        \end{minipage}
        \caption{Evaluation of different methods in the IWLS Contest 2020 benchmarks. The left (right) scatter shows the testing accuracy (the number of $AND$ gates).}
        \label{fig:5_1_IWLS_results}

\end{figure*}

%% file: Tables/5_1_IWLS_benchmarks_overview.tex
\begin{table}[h!]
    \centering
    \renewcommand{\arraystretch}{1.4}
    \tiny{}
        \begin{tabular}{|c|l|}
            \hline
            \footnotesize{\textbf{Case}} & \footnotesize{\textbf{Function Category}} \\\hline
            
            00-09 & 2 MSBs for $k$-bits adders, with $k \in \{16, 32, 64, 128, 256\}$ \\\hline

            10-19 & MSB of $k$-bits dividers and remainder circuits for $k \in \{16, 32, 64, 128, 256\}$ \\\hline

            20-29 & MSB and middle bit of $k$-bits multipliers for $k \in \{8, 16, 32, 64, 128\}$ \\\hline

            30-39 & $k$-bits comparators for $k \in \{10, 20, ..., 100\}$\\\hline

            40-49 & LSB and middle bit of $k$-bits square-rooters with $k \in \{16, 32, 64, 128, 256\}$\\\hline

            50-59 & 10 outputs of PicoJava designs with 16-200 inputs and roughly balanced on\&offset \\\hline
            
            60-69 & 10 outputs of MCNC i10 design with 16-200 inputs and roughly balanced on\&offset \\\hline
            
            70-79 & 5 other outputs from MCNC benchmarks + 5 symmetric functions of 16 inputs \\\hline
            
            80-89 & 10 binary classification problems from MNIST group comparison \\\hline
            
            90-99 & 10 binary classification problems from CIFAR-10 group comparison \\\hline
            
        \end{tabular}
        \caption{\label{tab:iwls20_benchmarks}An overview of the IWLS 2020 benchmarks.}
\end{table}

%% file: Tables/5_2_ISCAS85_benchmarks.tex
\begin{table}[htb]
    \centering
    \renewcommand{\arraystretch}{1.15}
    \setlength{\tabcolsep}{5pt}
    \footnotesize{}
    \begin{tabular}{ccc}\hline
    \small{\textbf{Circuits}} &  \small{\textbf{Circuit Function}} &  \small{\textbf{I / O}} \\\hline
            Add8u & 8-bit Adder & 16 / 9 \\
            Mul7u & 7-bit Multiplier & 14 / 14 \\
            C432 & 27-ch Interrupt Controller & 36 / 7 \\
            C499 & Single-Error-Corrector & 41 / 32 \\
            C880 & 8-bit ALU & 60 / 26 \\
             C1908 & 16-bit SEC/DED Circuit & 33 / 25 \\
            C3540 & 8-bit ALU (complicated) & 50 / 22 \\
            C6288 & 16-bit Multiplier & 32 / 32 \\ \hline
        \end{tabular}
        \caption{\label{tab:5_2_ISCAS_benchmarks}An overview of combinatorial circuits evaluated.}
\end{table}

%% file: Tables/5_2_results_C17.tex
\begin{table}[htb]
    \centering
    \renewcommand{\arraystretch}{1.2}
    \setlength{\tabcolsep}{1.2pt}
    \footnotesize{}
        \begin{tabular}{ccHccc}
            \hline
            \small{\textbf{Designs}} &  \small{\textbf{QoR($\%$)}} &  \small{\textbf{Gates}} &  \small{\textbf{Area($\mu m^{2}$)}} & \small{\textbf{Power($\mu W$)}} & \small{\textbf{Delay($ns$)}} \\\hline

            Original & $0$ & 2 $ANDNOT$, 2 $AND$, 2 $OR$ & 19.0 & 18.01 & 204.26 \\
            \hline
            
            BLASYS ($fd=1$) & $15.625$ & 1 $ANDNOT$, 1 $AND$, 1 $OR$ & 10.0 & 5.53 & 126.79 \\ \hline
            OPTDTALS ($\mdepth{} = 1$) & $25$ & / & 0.0 & 0.0 & 0.01 \\ 
            OPTDTALS ($\mdepth{} = 2$) & $12.5$ & 1 $MUX$, 1 $OR$ & 14.0 & 8.58 & 136.44 \\
           OPTDTALS ($\mdepth{} = 3$) & $6.25$ & 3 $ANDNOT$, 4 $MUX$, 1 $NAND$, 2 $NOR$, 1 $NOT$, 3 $OR$ & 20.0 & 18.10 & 124.42 \\
            OPTDTALS ($\mdepth{} = 4$) & $0$ & 1 $ANDNOT$, 6 $MUX$, 1 $NAND$, 1 $OR$ & 19.0 & 17.96 & 213.16 \\
            \hline
            
        \end{tabular}
        \caption{\label{tab:5_2_results_c17}Evaluations in the 
        C17 circuit.}
\end{table}

%% file: Tables/5_2_results_ISCAS_orig.tex
\begin{table}[htb]
    \centering
    \renewcommand{\arraystretch}{1.2}
    \setlength{\tabcolsep}{5pt}
    \footnotesize{}
        \begin{tabular}{cccc}
            \hline
            \small{\textbf{Circuits}} &  \small{\textbf{Area ($\mu m^{2}$)}} &  \small{\textbf{Power ($\mu W$)}} & \small{\textbf{Delay ($ns$)}} \\\hline
Add8u & 222.00 & 125.69 & 1009.32 \\
Mul7u & 1319.00 & 624.52 & 2182.10 \\
C432 & 529.00 & 130.48 & 1961.76 \\
C499 & 1270.00 & 1683.82 & 1503.03 \\
C880 & 1010.00 & 280.32 & 1750.13 \\
C1908 & 1243.00 & 586.49 & 1889.72 \\
C3540 & 2954.00 & 845.67 & 2714.61 \\
C6288 & 8493.00 & 100540.41 & 6693.01 \\
            \hline
            
        \end{tabular}
        \caption{\label{tab:5_2_ISCAS_orig}The summary of original circuit information.} 
\end{table}

%% file: Tables/5_2_results_ISCAS_approx.tex
\begin{table*}[htb]
    \centering
    \renewcommand{\arraystretch}{1.1}
    \setlength{\tabcolsep}{0.5pt}
    \scriptsize{}
        \begin{tabular}{cccccccccccccccccccccccccc} \hline
        \multirow{3}{*}{\small{\textbf{Cir}}} & 
        \multirow{3}{*}{\small{\textbf{Part}}} & 
        \multicolumn{12}{c}{\small{\textbf{BLASYS}}} & 
        \multicolumn{12}{c}{\small{\textbf{OPTDTALS-DL8.5}}}\\
        
        \cmidrule(lr){3-14}\cmidrule(lr){15-26} &
        & \multicolumn{4}{c}{{\textbf{$\mathbf{\leq 5\%} \; \errthreshold{}$}}} &
        \multicolumn{4}{c}{{\textbf{$\mathbf{\leq 10\%} \; \errthreshold{}$}}} &
        \multicolumn{4}{c}{{\textbf{$\mathbf{\leq 15\%} \; \errthreshold{}$}}} &
        \multicolumn{4}{c}{{\textbf{$\mathbf{\leq 5\%} \; \errthreshold{}$}}} &
        \multicolumn{4}{c}{{\textbf{$\mathbf{\leq 10\%} \; \errthreshold{}$}}} &
        \multicolumn{4}{c}{{\textbf{$\mathbf{\leq 15\%} \; \errthreshold{}$}}}\\
        
        \cmidrule(lr){3-6}\cmidrule(lr){7-10}\cmidrule(lr){11-14}\cmidrule(lr){15-18}\cmidrule(lr){19-22}\cmidrule(lr){23-26} & 
        & \textbf{A($\%$)} & \textbf{P($\%$)} & \textbf{D($\%$)} & \textbf{T(s)} &  \textbf{A($\%$)} & \textbf{P($\%$)} & \textbf{D($\%$)} & \textbf{T(s)} & \textbf{A($\%$)} & \textbf{P($\%$)} & \textbf{D($\%$)} & \textbf{T(s)} & 
        \textbf{A($\%$)} & \textbf{P($\%$)} & \textbf{D($\%$)} & \textbf{T(s)} & \textbf{A($\%$)} & \textbf{P($\%$)} & \textbf{D($\%$)} & \textbf{T(s)} & \textbf{A($\%$)} & \textbf{P($\%$)} & \textbf{D($\%$)} & \textbf{T(s)}\\ \hline

Add8u & 5 & - & - & - & - & \cellcolor{blue!40}91.0 & \cellcolor{blue!40}90.0 & \cellcolor{blue!40}87.0 & \cellcolor{blue!40}10.5 & - & - & - & - & \cellcolor{blue!40}98.2 & \cellcolor{blue!40}98.1 & \cellcolor{blue!40}94.5 & \cellcolor{blue!40}3.7 & - & - & - & - & \cellcolor{blue!40}76.6 & \cellcolor{blue!40}60.0 & \cellcolor{blue!40}40.0 & 11.7 \\

Mul7u & 31 & 96.4 & 97.4 & 98.8 & \cellcolor{blue!40}76.9 & \cellcolor{blue!40}88.5 & \cellcolor{blue!40}87.7 & 90.3 & \cellcolor{blue!40}93.8 & \cellcolor{blue!40}79.5 & \cellcolor{blue!40}79.1 & \cellcolor{blue!40}86.3 & \cellcolor{blue!40}222.9 & \cellcolor{blue!40}95.3 & \cellcolor{blue!40}97.0 & \cellcolor{blue!40}95.1 & 198.3 & 91.9 & 97.2 & \cellcolor{blue!40}90.0 & 253.0 & - & - & - & -\\

C432 & 7 & - & - & - & - & - & - & - & - & - & - & - & - & \cellcolor{blue!40}76.4 & \cellcolor{blue!40}82.3 & \cellcolor{blue!40}97.4 & \cellcolor{blue!40}261.8 & \cellcolor{blue!40}71.1 & \cellcolor{blue!40}72.9 & \cellcolor{blue!40}102.0 & \cellcolor{blue!40}399.7 & \cellcolor{blue!40}60.9 & \cellcolor{blue!40}51.5 & \cellcolor{blue!40}89.8 & \cellcolor{blue!40}482.8 \\

C499 & 40 & \cellcolor{blue!40}0.0 & \cellcolor{blue!40}0.0 & \cellcolor{blue!40}0.0 & \cellcolor{blue!40}1877.8 & \cellcolor{blue!40}0.0 & \cellcolor{blue!40}0.0 &\cellcolor{blue!40} 0.0 & \cellcolor{blue!40}2114.0 & \cellcolor{blue!40}0.0 & \cellcolor{blue!40}0.0 & \cellcolor{blue!40}0.0 & \cellcolor{blue!40}2228.1 & \cellcolor{blue!40}0.0 & \cellcolor{blue!40}0.0 & \cellcolor{blue!40}0.0 & 2896.1 & \cellcolor{blue!40}0.0 & \cellcolor{blue!40}0.0 & \cellcolor{blue!40}0.0 & 2909.3 & \cellcolor{blue!40}0.0 & \cellcolor{blue!40}0.0 & \cellcolor{blue!40}0.0 & 2915.8 \\

C880 & 35 & 72.5 & 77.1 & 77.9 & \cellcolor{blue!40}1085.4 & 47.9 & 44.6 & 55.7 & \cellcolor{blue!40}1540.8 & 25.6 & 26.3 & 45.7 & \cellcolor{blue!40}1827.2 & \cellcolor{blue!40}51.8 & \cellcolor{blue!40}49.1 & \cellcolor{blue!40}38.8 & 3013.4 & \cellcolor{blue!40}26.0 & \cellcolor{blue!40}31.4 & \cellcolor{blue!40}29.4 & 4192.1 & \cellcolor{blue!40}13.6 & \cellcolor{blue!40}15.3 & \cellcolor{blue!40}27.6 & 4502.9 \\

C1908 & 37 & 26.3 & \cellcolor{blue!40}22.3 & 45.0 & \cellcolor{blue!40}1731.0 & - & - & - & - & 5.1 & 4.8 & 27.2 & \cellcolor{blue!40}2127.0 & \cellcolor{blue!40}25.5 & 24.3 & \cellcolor{blue!40}41.6 & 2026.2 & \cellcolor{blue!40}14.0 & \cellcolor{blue!40}9.2 & \cellcolor{blue!40}38.7 & 2507.4 & \cellcolor{blue!40}0.0 & \cellcolor{blue!40}0.0 & \cellcolor{blue!40}0.0 & 2815.4 \\

C3540 & 113 & \cellcolor{blue!40}65.0 & \cellcolor{blue!40}78.5 & \cellcolor{blue!40}88.3 & \cellcolor{blue!40}23950.6 & 55.8 & 64.8 & 85.4 & \cellcolor{blue!40}25073.9 & 51.8 & 63.6 & 83.7 & \cellcolor{blue!40}25940.9 & 69.1 & 84.5 & 97.8 & 130811.3 & \cellcolor{blue!40}47.0 & \cellcolor{blue!40}63.5 & \cellcolor{blue!40}78.9 & 148032.8 & \cellcolor{blue!40}33.4 & \cellcolor{blue!40}40.7 & \cellcolor{blue!40}72.7 & 154308.1 \\

C6288 & 106 & \cellcolor{blue!40}91.8 & 302.6 & 100.6 & 48031.2 & \cellcolor{blue!40}89.0 & 301.7 & 101.3 & 55924.7 & 83.9 & 332.1 & 100.7 & 73402.3 & 92.7 & \cellcolor{blue!40}295.4 & \cellcolor{blue!40}100.4 & \cellcolor{blue!40}15737.1 & 89.4 & \cellcolor{blue!40}149.3 & \cellcolor{blue!40}99.3 & \cellcolor{blue!40}21383.8 & \cellcolor{blue!40}73.0 & \cellcolor{blue!40}59.6 & \cellcolor{blue!40}95.5 & \cellcolor{blue!40}48663.6 \\


\toprule[2pt]
\textbf{Avg} & 46.8 & 69.0 & 97.2 & 76.3 & \cellcolor{blue!40}12792.1 & 62.3 & 88.9 & 70.6 & \cellcolor{blue!40}12355.5 & 54.6 & 87.0 & 66.3 & \cellcolor{blue!40}15108.4 & \cellcolor{blue!40}63.6 & \cellcolor{blue!40}91.3 & \cellcolor{blue!40}70.7 & 19368.5 & \cellcolor{blue!40}54.7 & \cellcolor{blue!40}65.2 & \cellcolor{blue!40}66.6 & 22460.2 & \cellcolor{blue!40}43.7 & \cellcolor{blue!40}40.5 & \cellcolor{blue!40}52.0 & 26744.2 \\

\bottomrule[2pt]
        \end{tabular}
        \caption{\label{tab:5_2_ISCAS_approx}Evaluation of different methods in the large circuits with different increasing error thresholds.}
\end{table*}

%% file: paper_structure/6_conclusion.tex
\section{Conclusion}
\label{sec:conclusion}

We propose a novel Approximate Logic Synthesis methodology named 
{OPTDTALS via learning optimal decision trees in empirical accuracy.}
The guarantee of optimality in accuracy leads to a controllable trade-off between circuit complexity and accuracy via 
restricting the maximum depth for tree topology.
Our experimental studies show clear benefits of the proposed framework in providing more compact approximated designs with competitive accuracy compared with state-of-the-art heuristic ALS methods.

In the future,
it would be interesting to 
{extending our framework by applying different optimal ``white-box'' interpretable ML models, like decision diagrams, decision lists, etc.
Moreover, the effect of different partition algorithms applied for large circuits is also worth investigating.
}
Additionally, improving the scalability of our methodology is another crucial and urgent work.